\RequirePackage{fix-cm}
\documentclass[twocolumn]{svjour3} 

\smartqed
\usepackage{bm}
\usepackage[table]{xcolor}
\usepackage{graphicx}
\usepackage{newtxtext, newtxmath}
\usepackage{balance}
\usepackage{amsmath}
\usepackage[colorlinks=true, linkcolor=blue, citecolor=blue, urlcolor=blue]{hyperref}
\usepackage{xcolor}
\usepackage{latexsym}
\usepackage{makecell}
\usepackage{diagbox}
\usepackage{adjustbox}
\usepackage{caption}
\usepackage{subcaption}
\usepackage{soul}

\usepackage{booktabs}
\usepackage{multirow}
\usepackage{stfloats}
\usepackage{threeparttable}

\usepackage{url}
\usepackage{array}
\usepackage{ragged2e}
\usepackage{enumitem}
\usepackage{microtype}
\usepackage{pifont}
\usepackage{algorithm}
\usepackage{algorithmic}

\usepackage{enumerate}

\usepackage[round, sort]{natbib}

\begin{document}
\sloppy

\title{Open-Vocabulary Animal Keypoint Detection with Semantic-feature Matching}

\author{Hao Zhang \and
        Lumin Xu \and
        Shenqi Lai \and 
        Wenqi Shao \and
        Nanning Zheng$^{*}$ \and
        Ping Luo \and
        Yu Qiao \and
        Kaipeng Zhang$^{*}$ 
}

\institute{ $^{*}$ Corresponding Author\\
Hao Zhang \and Nanning Zheng \at
  National Key Laboratory of Human-Machine Hybrid Augmented Intelligence, National Engineering Research Center for Visual Information and Applications, and Institute of Artificial Intelligence and Robotics, Xi'an Jiaotong University, Xi'an, Shaanxi, 710049, China \\ 
  \email{zhanghao520@stu.xjtu.edu.cn,\\ nnzheng@mail.xjtu.edu.cn}
           \and
  Hao Zhang \and Kaipeng Zhang \and Wenqi Shao \and Yu Qiao \at
  Shanghai AI Laboratory, Shanghai, China \\
  \email{zhangkaipeng@pjlab.org.cn,\\
 shaowenqi@pjlab.orn.cn,\\
  qiaoyu@pjlab.org.cn}
           \and           
           Lumin Xu \at
  The Chinese University of Hong Kong, Hong Kong, China \\
  \email{luminxu@link.cuhk.edu.hk}
           \and
           Shenqi Lai \at
  Zhejiang University, Hangzhou, China \\
  \email{laishenqi@qq.com}
           \and
           Ping Luo  \at
  The University of Hong Kong, Hong Kong, China \\
  \email{pluo@cs.hku.edu}
}

\date{Received: 11 December 2023 / Accepted: 16 May 2024}

\maketitle

\begin{abstract}
Current image-based keypoint detection methods for animal (including human) bodies and faces are generally divided into fully supervised and few-shot class-agnostic approaches. The former typically relies on laborious and time-consuming manual annotations, posing considerable challenges in expanding keypoint detection to a broader range of keypoint categories and animal species. The latter, though less dependent on extensive manual input, still requires necessary support images with annotation for reference during testing.
To realize zero-shot keypoint detection without any prior annotation, we introduce the \textbf{O}pen-\textbf{V}ocabulary \textbf{K}eypoint \textbf{D}etection (OVKD) task, which is innovatively designed to use text prompts for identifying arbitrary keypoints across any species. 
In pursuit of this goal, we have developed a novel framework named Open-Vocabulary \textbf{K}eypoint \textbf{D}etection with \textbf{S}emantic-feature \textbf{M}atching (KDSM). This framework synergistically combines vision and language models, creating an interplay between language features and local keypoint visual features. KDSM enhances its capabilities by integrating \textbf{D}omain \textbf{D}istribution \textbf{M}atrix \textbf{M}atching (DDMM) and other special modules, such as the \textbf{V}ision-\textbf{K}eypoint \textbf{R}elational \textbf{A}wareness (VKRA) module, improving the framework’s generalizability and overall performance.
Our comprehensive experiments demonstrate that KDSM significantly outperforms the baseline in terms of performance and achieves remarkable success in the OVKD task. 
Impressively, our method, operating in a zero-shot fashion, still yields results comparable to state-of-the-art few-shot species class-agnostic keypoint detection methods.
Codes and data are available at https://github.com/zhanghao5201/KDSM.


\end{abstract}

\section{Introduction}\label{introduction}

\begin{figure*}[t]
\centering
  \includegraphics[width=0.9\textwidth]{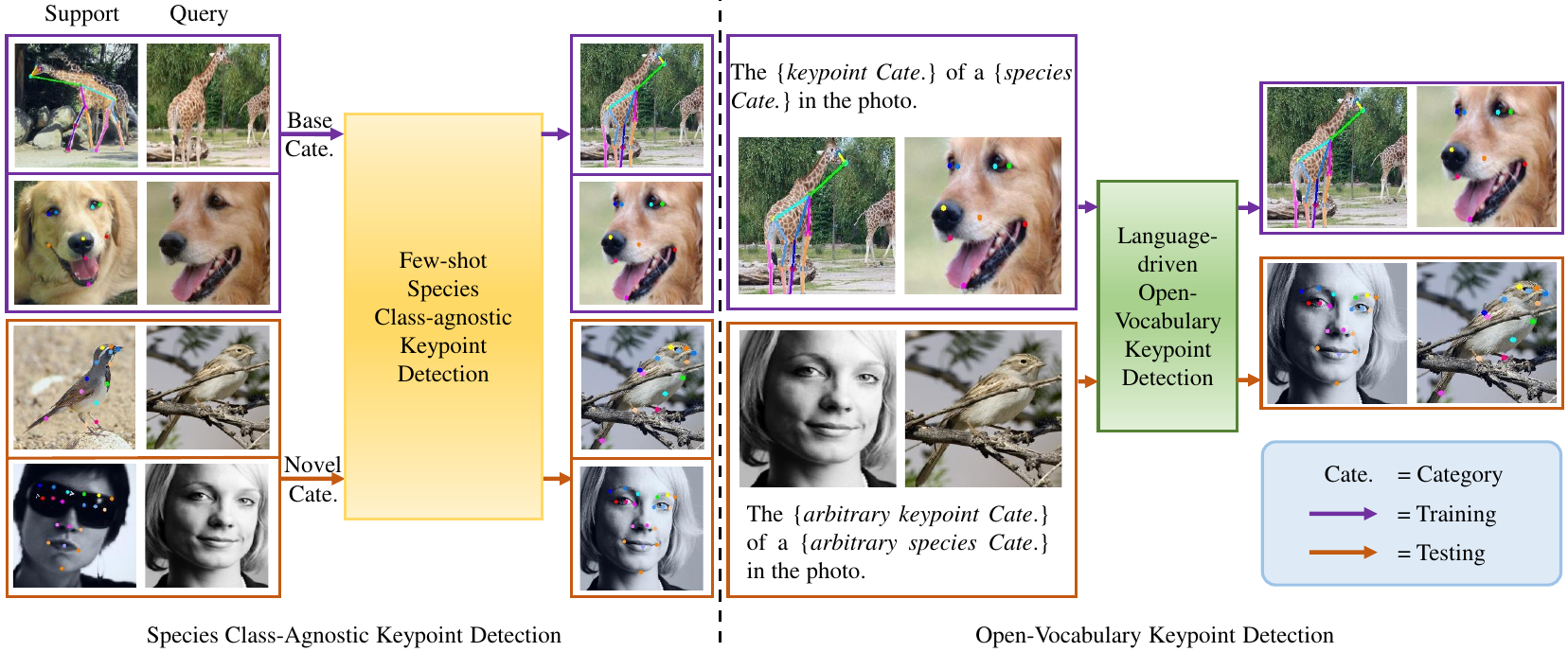}\\
{\footnotesize \qquad\qquad\qquad\qquad\qquad\qquad\quad\qquad(a)\hspace*{\fill}\qquad\qquad\qquad\qquad\qquad(b)\hspace*{\fill}}
  \caption{Few-shot Species Class-Agnostic Keypoint Detection vs. Language-driven Open-Vocabulary Keypoint Detection.
  (a) Current few-shot species class-agnostic keypoint detection needs support images for guidance during training and testing to detect keypoints in new species.
  (b) Language-driven OVKD aims to use text prompts that embed both $\{\textit{animal species}\}$ and $\{\textit{keypoint category}\}$ as semantic guidance to localize arbitrary keypoints of any species.
  }  
  \label{fig:intro}
\end{figure*}

Animal keypoint detection, a fundamental task in computer vision, is dedicated to identifying and localizing animals' keypoints within images. This task is pivotal for extensive analysis of animal (including human) bodies and faces. The accurate location of these keypoints plays a vital role in various applications, ranging from in-depth behavioral studies to automated monitoring systems, such as animal pose tracking~\citep{patel2023animal} and automatic assessment of animal pain~\citep{pessanha2022facial,feighelstein2022automated}.

Traditional keypoint detection methodologies have primarily centered around developing complex neural network architectures~\citep{andriluka20142d, newell2016stacked, Fang-ICCV-Rmpe-2017,wang2020deep,xu2022vitpose+,zhang2023scgnet,tu2023dual,zhang2024hfhrnet} and training them with datasets of annotated images to identify keypoints within specific species and keypoint categories. This strategy necessitates substantial manual labeling for each newly investigated species, often resulting in the creation of specialized datasets for these species~\citep{koestinger2011annotated,lin2014microsoft,brown2020language,labuguen2021macaquepose,khan2020animalweb}, a process known to be both time-consuming and labor-intensive.
For instance, compiling the AnimalWeb dataset~\citep{khan2020animalweb} required a substantial manual labeling effort totaling 6,833 man-hours from both experts and trained volunteers.
Despite such extensive manual efforts, the relatively limited availability and smaller size of animal keypoint datasets, compared to those for humans, present significant challenges in extending keypoint detection to new keypoint categories and animal species. The AnimalWeb dataset includes fewer than 239 annotations per species, in sharp contrast to the human-focused AFLW dataset~\citep{koestinger2011annotated}, which contains 25,993 annotations. Furthermore, some species in the AnimalWeb dataset are represented by only a single annotated image, making cross-species keypoint detection even more challenging, especially for species that lack annotations. Advanced few-shot species class-agnostic keypoint detection methods, as extensively detailed in studies like~\citep{xu2022pose, DBLP:conf/cvpr/0004HMH023}, represent progress in reducing the reliance on extensive manual annotations to adapt new keypoint categories and animal species.
As illustrated in Fig.~\ref{fig:intro} (a), these methods necessitate a small number of annotated support images for keypoint references during testing. In this paper, we further accomplish a more challenging task, which detects arbitrary keypoint in a zero-shot fashion without prior annotation during testing. 
Zero-shot keypoint detection could facilitate more convenient in-depth behavioral studies~\citep{patel2023animal} and the development of automated monitoring systems~\citep{pessanha2022facial,feighelstein2022automated} for new species and keypoint categories.

The potential of Vision-language models (VLMs)~\citep{radford2021learning, jia2021scaling} inspires our approach. 
VLMs have shown success in joint modeling of visual and text information, contributing to their exceptional zero-shot learning ability in various tasks, including object detection, semantic segmentation, video classification, and others~\citep{yao2022detclip,weng2023partcom,xu2023side}. However, there is a lack of research specifically addressing keypoint detection methods within this context.
Motivated by VLM advancements, we introduce the language-driven \textbf{O}pen-\textbf{V}ocabulary \textbf{K}eypoint \textbf{D}etection (OVKD) task (unless otherwise specified, OVKD always refers to language-driven OVKD). Specifically, OVKD is designed to identify a broad spectrum of $({\textit{animal species}}, {\textit{keypoint category}})$ pairs, including those not encompassed in the original training dataset. 
The term $\{\textit{keypoint category}\}$ refers to specific categories of keypoints, such as ``eyes" and ``nose." On the other hand, $\{\textit{animal species}\}$ represents a combination of the ``target keypoint detection task" and the corresponding animal species, encompassing categories like ``dog body," ``dog face," ``cat face," and ``cat body."
As shown in Fig.~\ref{fig:intro} (b), OVKD uses the image and text description of the keypoints to realize keypoint detection.

Building upon this concept, our initial strategy involves adopting a baseline framework (see Fig.~\ref{figbaseline}) that utilizes language models to obtain text embeddings for the descriptions of $({\textit{animal species}}, {\textit{keypoint category}})$ pairs. Then the baseline integrates the text embeddings with visual features using matrix multiplication and generates keypoint heatmaps. However, the limitation of this simple feature aggregation becomes evident in its lack of effective interaction between text and local visual features, hindering its ability to comprehend the local features of images and accurately localize specific keypoints. To address this, we emphasize the need for deeper interaction between text and local visual features of the image.

To overcome the limitations of the baseline framework in OVKD, we develop an advanced framework named Open-Vocabulary \textbf{K}eypoint \textbf{D}etection with \textbf{S}emantic-feature \textbf{M}atching (KDSM). 
KDSM introduces a \textbf{D}omain \textbf{D}istribution \textbf{M}atrix \textbf{M}atching (DDMM) technique and incorporates other special modules, such as a \textbf{V}ision-\textbf{K}eypoint \textbf{R}elational \textbf{A}wareness (VKRA) module, a keypoint encoder, a keypoint adapter, a vision head, and a vision adapter, among others.
The VKRA module uses attention blocks to enhance the interaction between text embeddings and local keypoint features. This facilitates a deeper exploration and understanding of the complex relationships between various local keypoint locations and text prompts during training.
Considering that the combinations of $({\textit{animal species}}, {\textit{keypoint category}})$ pairs are virtually infinite, it becomes impractical to construct a heatmap channel for every pair like fully supervised and few-shot species class-agnostic keypoint detection methods. Therefore, we propose DDMM, which utilizes clustering techniques to group the text features of $\{\textit{keypoint category}\}$. It allows semantically similar keypoint descriptions of different species to share a ground-truth heatmap channel representation during training. After grouping, the matching loss between text and heatmap features can be used to further align text features and keypoint visual features. 
During testing, DDMM assigns new text descriptions to specific groups, enabling the capability of zero-shot keypoint detection.

We conduct extensive experiments to evaluate the efficacy of our proposed method. The results emphatically demonstrate that our KDSM framework excels in OVKD, significantly surpassing the performance of the baseline framework. Notably, KDSM exhibits impressive zero-shot capabilities and comparable performance to the state-of-the-art few-shot species class-agnostic keypoint detection methods. The primary contributions of our research are summarized as follows:
\begin{itemize}
\item We introduce the task of OVKD, designed to utilize text prompts for detecting a diverse range of keypoint categories across different animal species in a zero-shot fashion.
\item We propose a pioneering approach, termed KDSM, to tackle the challenging OVKD task. DDMM technique and VKRA module are designed to model cross-species relationships and exchange vision-language information respectively.
\item Extensive experiments show that KDSM excels in OVKD, surpassing the baseline framework substantially. Despite operating in a zero-shot manner, KDSM achieves comparable results with state-of-the-art few-shot keypoint detection methods.

\end{itemize}

\section{Related Works}
Traditionally, the main research direction in keypoint detection has been fully supervised methods. This approach concentrated on improving keypoint detection accuracy via advancements in neural network architectures~\citep{andriluka20142d, newell2016stacked, Fang-ICCV-Rmpe-2017,wang2020deep,xu2022vitpose+,zhang2023scgnet,tu2023dual,zhang2024hfhrnet} and the development of new species datasets~\citep{koestinger2011annotated,lin2014microsoft,brown2020language,labuguen2021macaquepose}. However, these methods are confined to specific species or keypoint categories, limiting their adaptability to new types. Emerging few-shot category-agnostic keypoint detection techniques have started to address this, reducing the need for extensive annotations for novel species with a small number of annotated support images. We take this a step further by removing the necessity for image labeling and using language models to detect keypoints in a zero-fashion, open-vocabulary approach. 
In Section~\ref{2-1}, we will present the few-shot category-agnostic keypoint detection methods. Section~\ref{2-2} will introduce related works on open-vocabulary learning, and Section~\ref{2-3} will discuss the recent integration of language models with vision tasks.

\subsection{Advancements in Few-shot Species Class-agnostic Keypoint Detection}
\label{2-1}
A significant advancement in keypoint detection is the advent of few-shot species class-agnostic techniques~\citep{xu2022pose}, which can identify keypoints across various animal species without category-specific training. However, these techniques commonly rely on ``support images" during the training and testing phases. This reliance, characteristic of methods like MAML~\citep{finn2017model}, Fine-tune~\citep{nakamura2019revisiting}, FS-ULUS~\citep{lu2022few}, POMNet~\citep{xu2022pose}, and CapeFormer~\citep{DBLP:conf/cvpr/0004HMH023}, limits their applicability to new species or keypoints.

Specifically, POMNet~\citep{xu2022pose} initially proposed the few-shot species class-agnostic keypoint detection task and created the MP-100 expert dataset for it. CapeFormer~\citep{DBLP:conf/cvpr/0004HMH023} presents a two-stage framework incorporating techniques like a query-support refine encoder and a similarity-aware proposal generator for category-agnostic detection, shifting focus from heatmap prediction to keypoint position regression.
In contrast, our proposed OVKD task moves away from reliance on support images. OVKD leverages text prompts containing both $\{\textit{animal species}\}$ and $\{\textit{keypoint category}\}$, offering semantic guidance for detecting any keypoint in any species. This novel approach is aligned with zero-shot learning principles and marks a stride towards open-world animal body and facial keypoint detection.

\subsection{Exploring Open-Vocabulary Learning in Computer Vision}
\label{2-2}
Open-vocabulary learning, a burgeoning field in computer vision, has been explored in various tasks, including object detection~\citep{bangalath2022bridging,yao2022detclip}, semantic segmentation~\citep{li2022language,xu2023side}, 3D object recognition~\citep{weng2023partcom,zhu2023pointclip} and video classification~\citep{ni2022expanding, qian2022multimodal}. The advent of vision-language models like CLIP~\citep{radford2021learning} and ALIGN~\citep{jia2021scaling} has underscored their potential in tasks that require simultaneous processing of visual and text data, ideal for open-world learning scenarios.

While existing open-vocabulary learning research excels in image-level classification~\citep{zhu2023pointclip}, per-pixel classification~\citep{li2022language}, and mask classification~\citep{xu2023side}, keypoint detection poses a unique challenge. It demands not only a global understanding of the image but also precise localization of specific keypoints. To tackle this, we propose a novel technique called ``Domain Distribution Matrix Matching." This technique transforms keypoint detection into a task of aligning semantic-feature distributions from input text prompts with the detected heatmaps, thereby enhancing the accuracy and efficiency of the detection process.

\subsection{Leveraging Language Models for Vision Tasks}
\label{2-3}
Leveraging language models for vision tasks has ushered in a new era of methodologies that significantly enhance machines' understanding and interpretation of visual data. Some works~\citep{radford2021learning,jia2021scaling} utilize contrastive learning between language features and image features from vast collections of (image, text) pairs (e.g., 400 million in CLIP) to establish connections between language and visuals, which have marked the rapid development of using language models to aid in vision tasks. 
Specifically, open-vocabulary learning methods~\citep{bangalath2022bridging,xu2023side} employ pre-trained language and vision models to identify objects or scenes within images, showcasing extraordinary flexibility and adaptability. Furthermore, Large Vision-Language Models~\citep{chen2023sharegpt4v,lin2024moe} integrate CLIP's image encoder into language models, greatly facilitating tasks like Image Captioning and Visual Question Answering, among others. Additionally, language models gradually play a crucial role in basic visual tasks such as language-assisted image generation~\citep{rombach2022high,li2024blip} and multi-person pose estimation under occlusion~\citep{hu2023lamp}. Meanwhile, applying language models to vision tasks also presents numerous ethical and security challenges~\citep{zhang2024avibench}. Therefore, our work is dedicated to sensibly utilizing language models to assist in the open-vocabulary keypoint detection task.

\section{Method}\label{sec:method}

In this section, we begin by defining the \textbf{O}pen-\textbf{V}ocabulary \textbf{K}eypoint \textbf{D}etection (OVKD) task in Section~\ref{3-1}. We then present a baseline framework in Section~\ref{3-2}, which offers a straightforward solution to the task. In Section~\ref{3-3}, we introduce our proposed Open-Vocabulary \textbf{K}eypoint \textbf{D}etection with \textbf{S}emantic-feature \textbf{M}atching (KDSM) framework, outlining its unique design and capability.

\subsection{Problem Formulation: Open-Vocabulary Keypoint Detection}\label{3-1}

We introduce a novel task termed OVKD for animal (including human) body and face keypoint localization.
The goal of OVKD is to develop a framework capable of detecting arbitrary keypoints in images, even if the animal species or keypoint category is not present in the training data. 
The advancements in vision-language models such as CLIP~\citep{radford2021learning}, allow the keypoint detectors to take advantage of powerful language models to achieve language-driven OVKD.

For OVKD, text prompts are leveraged to guide the framework in understanding the semantic information and locating specific keypoints. 
Assuming we have a training set $\mathcal{D}_{train}$ and a test set $\mathcal{D}_{test}$, $\mathcal{D}_{train} = \{ (\mathit{\mathbf{I}}, T(s_i, k_j)_{j=1}^{\mathbb{K}_{s_i}}, G(s_i, k_j)_{j=1}^{\mathbb{K}_{s_i}}) \}_{i=1}^{\mathbb{S}}$, $\mathcal{D}_{test} = \{ (\mathit{\mathbf{I}}, T(s'_i, k'_j)_{j=1}^{\mathbb{K}'_{s'_i}}, G(s'_i, k'_j)_{j=1}^{\mathbb{K}'_{s'_i}}) \}_{i=1}^{\mathbb{S}'}$. Here, $\mathit{\mathbf{I}}$ represents images, $T(s_i, k_j)$ denotes the text prompts constructed based on species $s_i$ and keypoint category $k_j$, and $G(s_i, k_j)$ denotes the ground-truth heatmaps constructed based on the locations of the species $s_i$ and keypoint category $k_j$ in the images $\mathit{\mathbf{I}}$.
$\mathbb{S}$ and $\mathbb{K}_{s_i}$ represent the number of species and the number of keypoint categories of species $s_i$ in the training set, respectively, while $\mathbb{S}'$ and $\mathbb{K}'_{s'_i}$ represent the number of species and the number of keypoint categories of species $s'_i$ in the test set, respectively.  The test set includes $({\textit{animal species}}, {\textit{keypoint category}})$ pairs not covered in the training dataset, requiring the detector to identify arbitrary keypoints as per the text prompts.

\subsection{Baseline: A Simple Framework for OVKD} \label{3-2}
\begin{figure*}[t]
  \centering
  \includegraphics[width=0.9\textwidth]{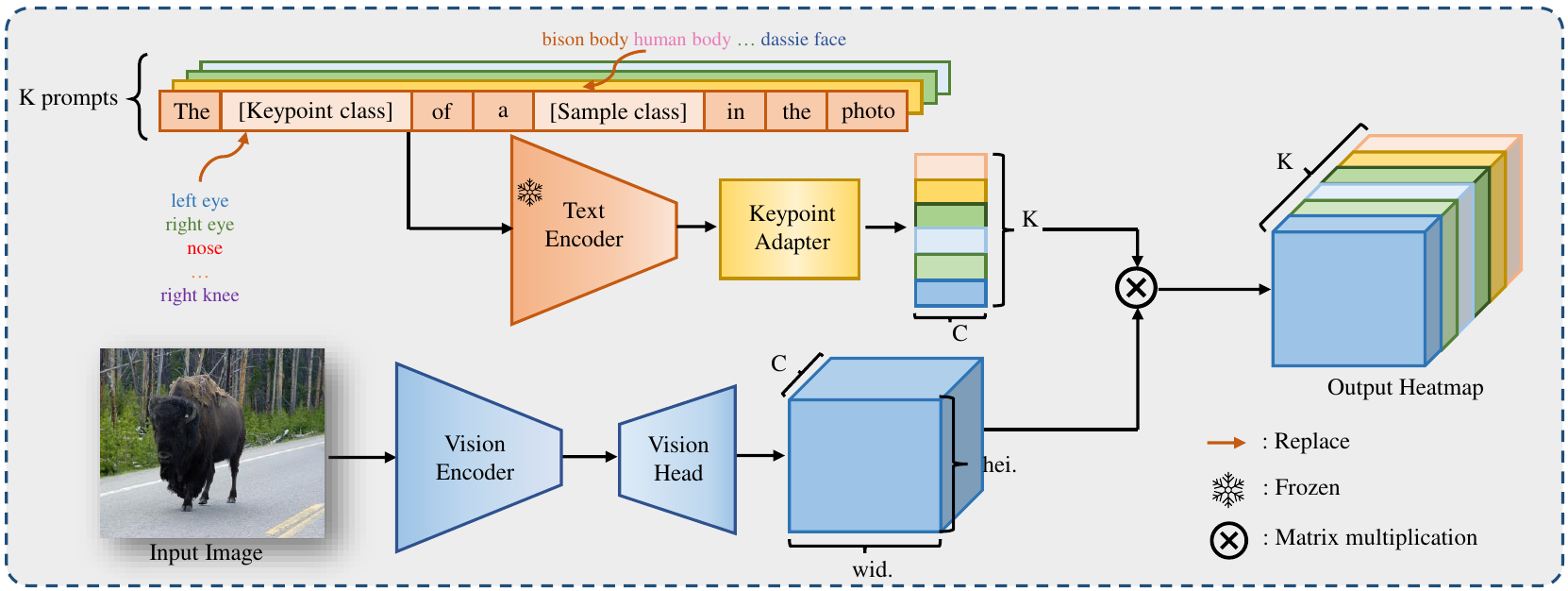}
  \caption{An overview of the baseline method for OVKD. The baseline comprises a $\mathrm{Vision\_Encoder}$, a $\mathrm{Text\_Encoder}$, a $\mathrm{Vision\_Head}$ and a $\mathrm{Keypoint\_Adapter}$.
  The $\mathrm{Keypoint\_Adapter}$ is applied to optimize the relevance of text features with the image features and produce the text feature with the shape of C$\times$K, where C and K represent the number of channel and text prompts, respectively. The $\mathrm{Vision\_Head}$ produces the visual feature with the shape of C$\times$hei.$\times$wid., where hei. and wid. represent the height and width, respectively.
  }
  \label{figbaseline}
\end{figure*}

To tackle the challenging OVKD task, we build a baseline framework that can predict arbitrary keypoint categories of any animal species as shown in Fig.~\ref{figbaseline}.
The baseline method constructs text prompts for the OVKD task and extracts text embedding using a $\mathrm{Text\_Encoder}$. The $\mathrm{Vision\_Encoder}$ is applied to extract visual features of the input image simultaneously. Then, the visual and text features are integrated by matrix multiplication to output heatmaps of keypoints defined by text prompts.

\textbf{Text Prompts Construction.} 
In this step, we utilize the template ``The $\{\textit{keypoint category}\}$ of a $\{\textit{animal species}\}$ in the photo." to assist language models in effectively grasping the task. For example, if ``giraffe body" is the animal species and ``neck" is the keypoint category, the prompt becomes: ``The ${\mathrm{neck}}$ of a ${\mathrm{giraffe\ body}}$ in the photo." This consistent template is applied across various animals and keypoints, with placeholders adjusted accordingly. Utilizing this template enables the language model to concentrate on the interplay between animal species and keypoints, facilitating smooth generalization to new species and keypoints within the open-vocabulary framework.
For the training and testing processes, the prompt construction is automatically generated using labeled datasets, i.e., $\{\textit{keypoint category}\}$ and $\{\textit{animal species}\}$ information.
If users are testing the system via an API, manual input of category information is indeed necessary, which is consistent with the open-vocabulary learning works mentioned in Section~\ref{2-2}.

\textbf{Text Feature Extraction.} Employing the pre-trained CLIP $\mathrm{Text\_Encoder}$~\citep{radford2021learning}, we process the preprocessed text prompts $T=\{T_1, T_2,..., T_K\}$ for an image with $K$ text prompts:
\begin{equation}
    \mathbf{T} = \mathrm{Keypoint\_Adapter}(\mathrm{Text\_Encoder}(T)),
\end{equation}
where $\mathrm{Text\_Encoder}(T)\in \mathbb{R}^{K\times C_0}$ represents the extracted text features. $\mathrm{Keypoint\_Adapter}$ is a two-layer Multi-layer Perceptron (MLP) used to refine these features and make them compatible with the image feature representations.
This refinement produces a semantic feature space $\mathbf{T}\in \mathbb{R}^{K\times C}$ (with $K = 100, C=64$ in our setup). 
$K$ represents the maximum number of keypoint categories for each species that can be handled, which can be adjusted as long as it is greater than the maximum number of keypoint categories across all species.
Due to the differences in the number of keypoints among different species, we insert $K-K_{valid}$ fixed invalid placeholder text features, where $K_{valid}$ denotes the number of valid text prompts. The text features of the invalid placeholders are derived from the prompt ``There is not the keypoint we are looking for."

\textbf{Vision Feature Extraction.}
Given an input image $I$, we train a $\mathrm{Vision\_Encoder}$ and a $\mathrm{Vision\_Head}$ to extract image features:
\begin{equation}
    \mathbf{V} = \mathrm{Vision\_Head}(\mathrm{Vision\_Encoder}(I)),
\end{equation}
where $\mathbf{V}\in \mathbb{R}^{C\times hei.\times wid.}$ ($hei.=64, wid.=64$ in our implementation) represents vision feature. We utilize ResNet~\citep{he2016deep} as the backbone of the $\mathrm{Vision\_Encoder}$, which is known to be effective in extracting hierarchical visual features from images.
The $\mathrm{Vision\_Head}$, inspired by SimpleBaseline~\citep{xiao2018simple}, is composed of three deconvolutional layers. These layers serve to upsample the low-resolution feature maps acquired from the image encoder, thereby successfully recovering spatial information and enabling accurate keypoint localization.

\textbf{Keypoint Heatmap Prediction.}
The objective of this framework is to predict keypoint localization by aggregating semantic text and spatial visual features.
To calculate the similarity between the text feature and pixel-level visual representation, the extracted features are combined through matrix multiplication:
\begin{equation}
    \mathbf{H} = \mathbf{T} \times \mathbf{V},
\end{equation}
where $\mathbf{H}\in \mathbb{R}^{K\times hei.\times wid.}$ denotes predicted heatmaps. 
The framework supports multiple text prompt inputs for detecting several keypoints simultaneously.
The model training is supervised using Mean Squared Error (MSE) loss between these predicted heatmaps $\mathbf{H}$ and the ground-truth heatmaps $\mathbf{G}\in \mathbb{R}^{K\times hei.\times wid.}$. 
In the construction of $\mathbf{G}$, each valid channel of the heatmap corresponds to a specific text prompt, which is in the form of ``The $\{\textit{keypoint category}\}$ of a $\{\textit{animal species}\}$ in the photo." We apply a 2D Gaussian with a standard deviation of 2 pixels, centered on the ground-truth location of the keypoint described by the prompt. The process of generating the Gaussian kernel is consistent with HRNet~\citep{wang2020deep} and POMNet~\citep{xu2022pose}.
Only the first $K_{valid}$ heatmaps are valid for $\mathbf{G}$, the other $(K-K_{valid})$ heatmaps are set to zero matrices. During the loss computation, only the first $K_{valid}$ channels of $\mathbf{G}$ and $\mathbf{H}$ are used.
During training, the $\mathrm{Text\_Encoder}$ remains frozen, while other parameters are trainable. 
The matrix multiplication operation conducts a transformation of visual features to the output heatmap spaces, driven by the semantic information contained in the text prompts.

\begin{figure*}[t]
  \centering
  \includegraphics[width=0.9\textwidth]{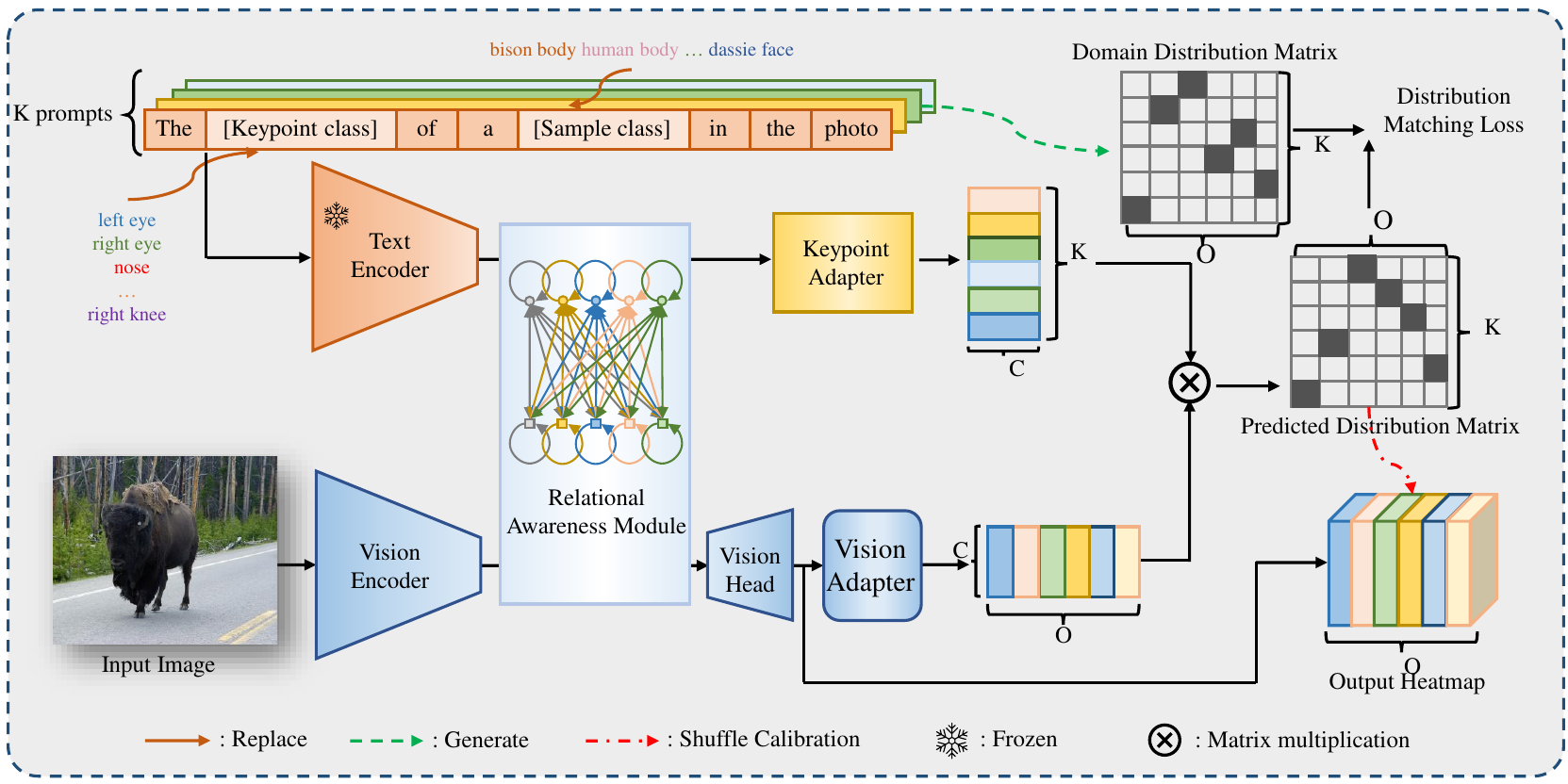}
  \caption{An overview of KDSM. KDSM comprises a $\mathrm{Vision\_Encoder}$, a $\mathrm{Text\_Encoder}$, a $\mathrm{Keypoint\_Adapter}$, a $\mathrm{Vision\_Adapter}$ and a $\mathrm{Vision\_Head}$ similar to the baseline. The vision-keypoint relational awareness module adjusts visual features according to their associations with keypoints. The $\mathrm{Vision\_Adapter}$ is employed to modify the feature shape so that it matches the text features' shape. Similarity is calculated between the adjusted features and text semantic features, resulting in a predicted distribution matrix. The predicted distribution matrix and the text domain distribution matrix are then utilized to compute matching loss.}
  \label{figmethod}
\end{figure*}

\subsection{Open-Vocabulary Keypoint Detection with Semantic-feature Matching} \label{3-3}

In this section, we propose a novel framework, namely KDSM, to address the limitations of the baseline OVKD framework. 
The baseline framework just uses simple feature aggregation, which fails to effectively capture the intricate relationship between text and local visual features and establish clear connections between them, leading to less than optimal keypoint detection.
Therefore, KDSM proposes \textbf{D}omain \textbf{D}istribution \textbf{M}atrix \textbf{M}atching (DDMM) and adopts some special modules to address the above problems, such as a \textbf{V}ision-\textbf{K}eypoint \textbf{R}elational \textbf{A}wareness (VKRA) module, a keypoint encoder, a keypoint adapter, a vision head, and a vision adapter, among others.

As depicted in Fig.~\ref{figmethod}, KDSM initially constructs text prompts and extracts text features similarly to the baseline approach. However, it then employs the VKRA module to facilitate a deeper exploration and understanding of the complex relationships between various local keypoint locations and text prompts during training.
Finally, DDMM is proposed to capture cross-species keypoint-level relationships to further enhance the generalization ability of KDSM. Notably, KDSM supports multiple text prompt inputs for detecting several keypoints simultaneously.

\textbf{Vision-Keypoint Relational Awareness Module.}
Within our framework, the VKRA module, incorporating a series of Transformer blocks inspired by~\citep{pan2020x}, is an essential design. It comprises two main components: Self-Attention~\citep{vaswani2017attention} ($\mathrm{Self\_Attn.}$) and Cross-Attention~\citep{carion2020end} ($\mathrm{Cross\_Attn.}$). The self-attention layers are designed to enhance the interaction among text embeddings of the given sample. They amalgamate keypoint features as follows:
\begin{equation}
\mathbf{Y}_t = \mathrm{Self\_Attn.}(\mathrm{Text\_Encoder}(\mathbf{T})).
\end{equation}
The refined keypoint features $\mathbf{Y}_t$ elucidate the relationships of text semantic concepts among the keypoints of a specific species. 

The cross-attention layers use the output features from the $\mathrm{Vision\_Encoder}$ as the query, while the refined features $\mathbf{Y}_t$ serve as the key and value. This mechanism facilitates interaction between the context-aware visual features $\mathrm{Vision\_Encoder}(I)$ and the refined features $\mathbf{Y}_t$ to enhance the vision representation:
\begin{equation}
\widetilde{\mathbf{V}} = \mathrm{Cross\_Attn.}(\mathrm{Vision\_Encoder}(I), \mathbf{Y}_t)
\end{equation}
The updated visual features $\widetilde{\mathbf{V}}$ effectively capture the relationships between local visual features and keypoint text features, bridging the gap between vision representation and the keypoint.

\textbf{Domain Distribution Matrix Matching.}
To model cross-species keypoint-level relationships, we propose a domain distribution matrix that links keypoint categories to corresponding output heatmaps. Assuming we have 78 species and 15 keypoint categories per species, this leads to 1170 hypothetical $({\textit{animal species}}, {\textit{keypoint category}})$ combinations. 
Directly representing each combination with a unique heatmap channel to model the above relationships is impractical. 
There exists cross-species commonality at the keypoint category level for OVKD since the keypoints of different animals may be similar. The similarity could be grasped during training by dividing all the keypoint categories into several groups and learning keypoint categories in the same group together.
Therefore, we opt to represent multiple keypoint categories of different species using a single channel of ground-truth heatmaps.
By grouping all keypoint categories and learning them collectively within these groups (where each group corresponds to one heatmap channel), we enhance the efficiency of the training process and avoid unnecessary computational expenditure.
Notably, only keypoint categories of different species that are clustered into the same group will share the same ground-truth heatmap channel, and all keypoint categories of the same species are clustered into different groups (ground-truth heatmap channels) in our setting.
During testing, a new $({\textit{animal species}}, {\textit{keypoint category}})$ combination is assigned to one of the predefined groups based on the predicted distribution matrix. The heatmap representation of the selected group is then utilized to detect keypoints for that specific combination.
Consequently, domain distribution matrix matching plays a crucial role in enhancing the prediction of new keypoint categories across various species. 

Specifically, we apply K-means clustering to all training set keypoint categories, dividing them into $O$ groups based on text embeddings generated by the $\mathrm{Text\_Encoder}$ from $\{\textit{keypoint category}\}$ terms, we set that all $\{\textit{keypoint category}\}$ of the same species must belong to different groups in the clustering process. We then pre-compute a binary domain distribution matrix $\mathbf{D} \in \mathbb{R}^{K\times O}$ (setting $O=100$) for each training sample, based on its keypoint categories. Here, $K$ is a constant no smaller than any sample's maximum keypoint count. We set $\mathbf{D}_{ij} = 1$ when the $i$-th keypoint falls into the $j$-th group. If a sample's keypoint count $K'$ is less than $K$, $\mathbf{D}_{ij} = 0$ for $i \in [K' + 1, K]$ and $j \in [1, O]$.

To learn group selection, we predict the distribution matrix. First, the updated visual features $\widetilde{\mathbf{V}}$ are merged with the original features to strengthen visual representation. These features pass through the $\mathrm{Vision\_Head}$, identical to the baseline, to generate heatmaps $\mathbf{H}' \in \mathbb{R}^{O\times hei.\times wid.}$. The $\mathrm{Vision\_Adapter}$ then generates the visual features $\mathbf{V}'$ from these heatmaps, and the $\mathrm{Keypoint\_Adapter}$ adapts the original text embeddings and generates $\mathbf{T}'$. Finally, we measure the similarity between the adjusted visual features $\mathbf{V}' \in \mathbb{R}^{C \times O}$ and the adjusted text embeddings $\mathbf{T}' \in \mathbb{R}^{K \times C}$ to create a predicted distribution matrix $\mathbf{P} \in \mathbb{R}^{K \times O}$:
\begin{equation}
\mathbf{P} = \mathbf{T}' \times \mathbf{V}'.
\end{equation}

\textbf{Loss Function.}
The matching loss $L_{match}$, computed as the cross-entropy loss between the predicted distribution matrix $\mathbf{P}$ and the domain distribution matrix $\mathbf{D}$, aims to align keypoint categories with heatmap channels:
\begin{equation}
L_{match} = -\sum_{i=1}^{K}\sum_{j=1}^{O} \mathbf{D}_{ij} \log{\mathbf{P}_{ij}}.
\end{equation}

During training, we utilize the annotated domain distribution matrix $\mathbf{D}$ to determine the $j$-th heatmap position for the $i$-th prompt, addressing the mismatch between predicted heatmap ordering and prompt sequencing. During testing, alignment is achieved using the predicted domain distribution matrix $\mathbf{P}$. Subsequently, heatmaps $\mathbf{H}'$ produced by $\mathrm{Vision\_Head}$ are reorganized across channels based on $\mathbf{D}$ during training or $\mathbf{P}$ during testing to ensure correct alignment with their respective prompts. This reordering involves identifying the index $o$ of the element $1$ in the $i^{th}$ row of $\mathbf{D}$ or $\mathbf{P}$, signifying the $o^{th}$ channel of $\mathbf{H}'$ as matching the $i^{th}$ prompt. PyTorch functions like ``torch.index\_select" facilitate this reordering process.
The reordered heatmaps $\mathbf{H} \in \mathbb{R}^{O \times hei. \times wid.}$ are then evaluated against the ground-truth heatmaps $\mathbf{G} \in \mathbb{R}^{O \times hei. \times wid.}$ using the Mean Squared Error (MSE) loss. The initial $K_{valid}$ channels of $\mathbf{G}$ correspond to keypoint locations identified by the $K_{valid}$ text prompts, while the remaining $O - K_{valid}$ channels are treated as invalid zero matrices. The overall training loss for KDSM is defined as:
\begin{equation}
L_{total} = \alpha L_{match} + \beta MSE(\mathbf{H},\mathbf{G}),
\end{equation}
where $\alpha$ and $\beta$ are the balance weights, and they are set to $1e^{-6}$ and $1$ unless otherwise specified. 
The process of generating $\mathbf{G}\in \mathbb{R}^{O\times hei.\times wid.}$ ($O=100, hei.=64, wid.=64$ in our implementation.) mirrors that of the baseline in Section~\ref{3-2}, except for the total number of channels. In KDSM, the total number of channels $O$ in $\mathbf{G}$ is predefined as the number of clusterings, distinct from $K$ in the baseline, which represents the number of prompts.

\textbf{Inference Process.}
During the inference phase, when presented with an input image and corresponding text prompts, KDSM replicates its training methodology to estimate the keypoint heatmaps and the predicted distribution matrix. This process involves a detailed analysis for each keypoint category $k$. Specifically, we search for the maximum value in the $k$-th row of the predicted distribution matrix $\mathbf{P}$, which identifies the index of the corresponding heatmap channel for that particular keypoint.

Once the indexes are determined, the heatmaps are carefully reordered and calibrated to align with these indexes, thus serving as the final prediction results. This step is crucial in ensuring the accuracy of our keypoint localization. Subsequently, the keypoint localization is precisely decoded as the coordinates that correspond to the highest scores within these reordered heatmaps.

In our experiment, we simply use the \textit{maximum value indexing} as mentioned earlier. Our statistical analysis showed no samples of different keypoints corresponding to the same heatmap. However, variations in the test set might result in overlapping assignments, which motivates us to develop a fast-indexing algorithm (Algorithm~\ref{alg:assign_heatmaps}). Algorithm 1 does not affect the accuracy of our experimental results. Furthermore, due to semantic similarities and pose variations, it is normal and acceptable for multiple keypoints to occasionally map to the same heatmap. Therefore, Algorithm~\ref{alg:assign_heatmaps} is offered as an optional solution, allowing users to choose based on their specific requirements.

\begin{algorithm}
\caption{Assign Heatmaps to Keypoints Based on the Predicted Domain Distribution Matrix During Inference}
\label{alg:assign_heatmaps}
\begin{algorithmic}[1]
\REQUIRE Predicted Domain Distribution Matrix $\mathbf{P}$ of size $K \times O$, where $K$ is the number of keypoints and $O$ is the number of heatmaps.
\ENSURE $L$: A list of heatmap indices assigned to each keypoint.

\STATE Initialize a priority queue $Q$.
\STATE Initialize an empty set of assigned heatmaps $A_O$.
\STATE Initialize an empty set of assigned keypoints $A_K$.
\STATE Initialize the assignments list $L$ with $-1$ for each keypoint.

\FOR{$i = 1$ to $K$}
    \FOR{$j = 1$ to $O$}
        \STATE Add $(\mathbf{P}[i, j], i, j)$ to the priority queue $Q$.
    \ENDFOR
\ENDFOR

\WHILE{not $Q$.isEmpty() AND $|A_K| < K$}
    \STATE Extract the maximum score entry $(score, k, o)$ from $Q$.
    \IF{$k \notin A_K$ AND $o \notin A_O$}
        \STATE Assign heatmap $o$ to keypoint $k$: $L[k] = o$.
        \STATE Add $k$ to the set of assigned keypoints $A_K$.
        \STATE Add $o$ to the set of assigned heatmaps $A_O$.
    \ENDIF
\ENDWHILE

\RETURN $L$
\end{algorithmic}
\end{algorithm}

\section{Experiments}\label{sec:experiments}

\subsection{Open-Vocabulary Evaluation Protocol}

\begin{table*}[t]
  \centering
  \caption{Comparisons with the baseline framework on the MP-78 dataset for Setting A with PCK@0.2, PCK@0.05 and NME. $\uparrow$ indicates higher is better, while $\downarrow$ indicates lower is better.}
    \setlength{\tabcolsep}{5.58mm}{
    \footnotesize
    \begin{tabular}{c|c|ccccc|c}
    \hline
    \rowcolor[HTML]{ECF4FF} Metric&Framework & Split1 & Split2 & Split3 & Split4 & Split5 &Mean Metric \\    
    \hline    
    \rowcolor[HTML]{EFEFEF} 
    \hline
     &Baseline& 42.02&44.00&42.55&43.80&42.26&42.93 \\
     \rowcolor[HTML]{EFEFEF} 
      \multirow{-2}{*}{\cellcolor[HTML]{EFEFEF}PCK@0.2 $\uparrow$}&KDSM & \textbf{87.93}&\textbf{88.50}&\textbf{87.64}&\textbf{88.28}&\textbf{88.82}&\textbf{88.23} \\ 
      \hline
      \rowcolor[HTML]{ECF4FF}      
      \hline
    &  Baseline&  11.08 &  11.44& 10.35& 14.98 & 11.80 & 11.93 \\
    \rowcolor[HTML]{ECF4FF}
     \multirow{-2}{*}{\cellcolor[HTML]{ECF4FF}PCK@0.05 $\uparrow$} &KDSM & \textbf{62.80}&\textbf{63.11}&\textbf{61.91}&\textbf{62.00}&\textbf{61.91}&\textbf{62.35}  \\
    \hline 
    \rowcolor[HTML]{EFEFEF} 
    \hline
     & Baseline    &29.96     & 29.18     & 30.54     & 29.31    & 29.60     &    29.72 \\
\rowcolor[HTML]{EFEFEF} 
 \multirow{-2}{*}{\cellcolor[HTML]{EFEFEF}NME $\downarrow$}& KDSM & \textbf{8.20}   & \textbf{7.55}   & \textbf{8.23}   & \textbf{7.81}   & \textbf{7.84}   &   \textbf{7.93}     \\
    \hline    
    \end{tabular}%
    }
  \label{tabsy1}%
\end{table*}%
\begin{table*}[t]
  \centering
  \caption{Comparisons on MP-78 dataset for Setting B. KDSM notably demonstrates comparable performance on par with other few-shot species class-agnostic keypoint detection approaches. We use different colors to show the \textcolor{magenta}{\textbf{best}} and \textcolor{green}{\textbf{second-best}} results respectively.}
    \setlength{\tabcolsep}{2.2mm}
    \footnotesize
    {\begin{tabular}{c|c|c|ccccc|c}%cyan、magenta
    \hline
\cellcolor[HTML]{ECF4FF} Metric & \cellcolor[HTML]{ECF4FF}Framework   & \cellcolor[HTML]{ECF4FF}Shot setting & \cellcolor[HTML]{ECF4FF}Split1    & \cellcolor[HTML]{ECF4FF}Split2    & \cellcolor[HTML]{ECF4FF}Split3    & \cellcolor[HTML]{ECF4FF}Split4    & \cellcolor[HTML]{ECF4FF}Split5    & \cellcolor[HTML]{ECF4FF}Mean Metric\\ \hline
\rowcolor[HTML]{EFEFEF} 
\hline
\cellcolor[HTML]{EFEFEF}& MAML~\citep{finn2017model}& 5-shot& 76.37     & 75.53     & 71.15     & 69.46     & 67.55     & 72.01     \\
\rowcolor[HTML]{EFEFEF} 
\cellcolor[HTML]{EFEFEF}& Fine-tune~\citep{nakamura2019revisiting}     & 5-shot& 77.81     & 76.51     & 72.55     & 71.09     & 69.85     & 73.56     \\
\rowcolor[HTML]{EFEFEF} 
\cellcolor[HTML]{EFEFEF}& FS-ULUS~\citep{lu2022few} & 5-shot& 78.34     & 79.67     & 76.89     & 81.52     & 75.23     & 78.33     \\
\rowcolor[HTML]{EFEFEF} 
\cellcolor[HTML]{EFEFEF}& POMNet~\citep{xu2022pose} & 5-shot& 81.25     & 86.44     & 81.01     & 86.93     & 78.68     & 82.86     \\
\rowcolor[HTML]{EFEFEF} 
\cellcolor[HTML]{EFEFEF}& CapeFormer~\citep{DBLP:conf/cvpr/0004HMH023} & 5-shot& \textcolor{magenta}{\textbf{91.01}} & \textcolor{magenta}{\textbf{90.95}} & \textcolor{magenta}{\textbf{87.90}} & \textcolor{magenta}{\textbf{91.90}} & \textcolor{magenta}{\textbf{87.23}} & \textcolor{magenta}{\textbf{89.80}} \\
\rowcolor[HTML]{EFEFEF} 
\cellcolor[HTML]{EFEFEF}& MAML~\citep{finn2017model}& 1-shot& 75.11     & 74.31     & 69.80     & 68.22     & 67.44     & 70.98     \\
\rowcolor[HTML]{EFEFEF} 
\cellcolor[HTML]{EFEFEF}& Fine-tune~\citep{nakamura2019revisiting}     & 1-shot& 76.65     & 76.41     & 71.37     & 69.97     & 69.36     & 72.75     \\
\rowcolor[HTML]{EFEFEF} 
\cellcolor[HTML]{EFEFEF}& FS-ULUS~\citep{lu2022few} & 1-shot& 73.69     & 70.65     & 63.97     & 71.14     & 63.65     & 68.62     \\
\rowcolor[HTML]{EFEFEF} 
\cellcolor[HTML]{EFEFEF}& POMNet~\citep{xu2022pose} & 1-shot& 73.07     & 77.89     & 71.79     & 78.76     & 70.26     & 74.35     \\
\rowcolor[HTML]{EFEFEF} 
\cellcolor[HTML]{EFEFEF}& CapeFormer~\citep{DBLP:conf/cvpr/0004HMH023} & 1-shot& 85.41     & 88.39     & 83.53     & 85.74     & 80.04     & 84.62     \\ \cline{2-9}
\rowcolor[HTML]{EFEFEF} 
\cellcolor[HTML]{EFEFEF}& Baseline    & zero-shot         & 56.06     & 55.36     & 54.35     & 53.07     & 50.66     & 53.90     \\
\rowcolor[HTML]{EFEFEF} 
\multirow{-12}{*}{\cellcolor[HTML]{EFEFEF}PCK@0.2 $\uparrow$}     & KDSM & zero-shot         & \textcolor{green}{\textbf{85.48}}   & \textcolor{green}{\textbf{89.45}}   & \textcolor{green}{\textbf{84.29}}   & \textcolor{green}{\textbf{86.25}}   & \textcolor{green}{\textbf{81.17}}   & \textcolor{green}{\textbf{85.33}}   \\ \hline 
\rowcolor[HTML]{ECF4FF} 
\hline
\cellcolor[HTML]{ECF4FF}& CapeFormer~\citep{DBLP:conf/cvpr/0004HMH023} & 5-shot& \textcolor{green}{\textbf{46.90}}   & \textcolor{green}{\textbf{51.90}}   & \textcolor{green}{\textbf{44.45}}   & \textcolor{green}{\textbf{52.30}}   & \textcolor{green}{\textbf{39.21}}   & \textcolor{green}{\textbf{46.95}} \\
\rowcolor[HTML]{ECF4FF} 
\cellcolor[HTML]{ECF4FF}& CapeFormer~\citep{DBLP:conf/cvpr/0004HMH023} & 1-shot& 40.59     & 44.13     & 35.59     & 42.34     & 33.00     & 39.13     \\ \cline{2-9} 
\rowcolor[HTML]{ECF4FF} 
\cellcolor[HTML]{ECF4FF}& Baseline    & zero-shot         & 32.40     & 32.20     & 29.37     & 30.67     & 27.13   & 30.35    \\
\rowcolor[HTML]{ECF4FF} 
\multirow{-4}{*}{\cellcolor[HTML]{ECF4FF}PCK@0.05 $\uparrow$}     & KDSM & zero-shot  & \textcolor{magenta}{\textbf{60.26}} & \textcolor{magenta}{\textbf{61.17}} & \textcolor{magenta}{\textbf{55.08}} & \textcolor{magenta}{\textbf{55.96}} & \textcolor{magenta}{\textbf{48.53}} & \textcolor{magenta}{\textbf{56.20}}  \\ \hline
\rowcolor[HTML]{EFEFEF} 
\hline
\multicolumn{1}{l|}{\cellcolor[HTML]{EFEFEF}}   & CapeFormer~\citep{DBLP:conf/cvpr/0004HMH023} & 5-shot& \textcolor{magenta}{\textbf{8.63}} & \textcolor{magenta}{\textbf{7.81}} & \textcolor{magenta}{\textbf{9.85}} & \textcolor{magenta}{\textbf{8.02}} & \textcolor{magenta}{\textbf{10.15}} &  \textcolor{magenta}{\textbf{8.89}}         \\
\rowcolor[HTML]{EFEFEF} 
\multicolumn{1}{l|}{\cellcolor[HTML]{EFEFEF}}   & CapeFormer~\citep{DBLP:conf/cvpr/0004HMH023} & 1-shot& 10.84     & 9.58     & 11.77     & 10.65     & 13.16     &    11.20       \\ \cline{2-9}
\rowcolor[HTML]{EFEFEF} 
\multicolumn{1}{l|}{\cellcolor[HTML]{EFEFEF}}   & Baseline    & zero-shot         & 23.78     & 25.21     & 25.62     & 25.92    &26.30    &    25.37      \\
\rowcolor[HTML]{EFEFEF} 
\multicolumn{1}{l|}{\multirow{-4}{*}{\cellcolor[HTML]{EFEFEF}NME $\downarrow$}} & KDSM & zero-shot         & \textcolor{green}{\textbf{9.71}}   & \textcolor{green}{\textbf{8.04}}   & \textcolor{green}{\textbf{10.96}}   & \textcolor{green}{\textbf{9.58}}   & \textcolor{green}{\textbf{12.16}}   &   \textcolor{green}{\textbf{10.09}}         \\ \hline

    \end{tabular}%
    }
  \label{tabsy2}%
\end{table*}%

\textbf{Dataset Split.}
MP-100~\citep{xu2022pose} is introduced for category-agnostic pose estimation, which contains over 20K instances covering 100 sub-categories and 8 super-categories (human hand, human face, animal body, animal face, clothes, furniture, and vehicle).
However, some of the keypoint categories in MP-100, such as those for clothes and furniture, lack practical semantic information and are not suitable for language-driven OVKD.
Thus, we selected a subset of 78 animal categories (including humans) with keypoint annotations that have specific, meaningful semantic information.
We call this subset ``MP-78", including COCO~\citep{lin2014microsoft}, AFLW~\citep{koestinger2011annotated}, OneHand10K~\citep{wang2018mask}, AP-10K~\citep{yu2021ap}, Desert Locust~\citep{graving2019deepposekit}, MascaquePose~\citep{labuguen2021macaquepose}, Vinegar Fly~\citep{pereira2019fast}, AnimalWeb~\citep{khan2020animalweb}, CUB-200~\citep{welinder2010caltech}.

MP-78 encompasses more than 14,000 images accompanied by 15,000 annotations. For keypoint types possessing semantic meaning, albeit lacking a precise definition or description, we employ ChatGPT to query and acquire the names of these keypoints.
For example, we use a query like ``How to anatomically describe the second joint of the index finger?" to obtain the name of a specific keypoint. All these queries are performed manually, and then we build the dataset MP-78.

It is essential to clarify that in this paper, $\{\textit{animal species}\}$ refers to a combination of ``target keypoint detection task $+$ animal species." For instance, the face and body of a dog are categorized as two distinct $\{\textit{animal species}\}$ entities (i.e., ``dog face" and ``dog body"), based on the specific keypoint detection task. This means that our definition of species extends beyond mere biological classification, encapsulating task-specific categories within each animal.

To evaluate the generalization ability of OVKD to different keypoint categories and animal species, we design two settings, that is ``Setting A: Diverse Keypoint Categories" for new $\{\textit{keypoint category}\}$, and ``Setting B: Varied Animal Species" for new $\{\textit{animal species}\}$ like~\citep{xu2022pose}.
All zero-shot settings strictly fall under ``transductive generalized zero-shot learning~\citep{pourpanah2022review}".

In Setting A, we divide the keypoint categories associated with each of the 78 species into two parts: seen $\{\textit{keypoint category}\}$ and unseen $\{\textit{keypoint category}\}$. During training, we only used the seen categories, while the unseen categories were reserved for testing. 
For fair evaluation, we randomly split seen $\{\textit{keypoint category}\}$ for each species to form seen $\{\textit{keypoint category}\}$ sets. We form five different train/test sets splits.

In Setting B, MP-78 is split into train/test sets, with 66 $\{\textit{animal species}\}$ for training, and 12 $\{\textit{animal species}\}$ for testing. 
To ensure the generalization ability of the framework, we evaluate the framework performance on five splits like~\citep{xu2022pose}, where each $\{\textit{animal species}\}$ is treated as a novel one on different splits to avoid $\{\textit{animal species}\}$ bias.

\begin{table*}[t]
  \centering
  \caption{Impact of hyperparameter settings on the performance (PCK0.2) of KDSM in Setting A for the OVKD task. 
}
    \setlength{\tabcolsep}{6.2mm}{\begin{tabular}{c|c|ccccc|c}
    \hline
    \rowcolor[HTML]{ECF4FF}
    $\alpha$ & $\beta$ &  Split1 &Split2 &Split3 &Split4& Split5& Mean (PCK@0.2 $\uparrow$) \\
    \hline \rowcolor[HTML]{EFEFEF} 
    1 & 1& 13.57 &13.22 &13.26&12.80&13.78&13.32 \\ \rowcolor[HTML]{EFEFEF} 
    1$\times 1e^{-1}$ & 1 &42.87&31.06&32.34&14.16&31.32&30.35\\ \rowcolor[HTML]{EFEFEF} 
    1$\times 1e^{-3}$ & 1&79.02&71.35&76.68&79.79&74.67&76.30 \\ \rowcolor[HTML]{EFEFEF} 
    1$\times 1e^{-4}$ & 1&83.99&79.96&87.50&87.32&85.86&84.93 \\   \rowcolor[HTML]{EFEFEF}   
    1$\times 1e^{-6}$ &1 & 87.93 &88.50 &87.64 &88.28 &88.82&\textbf{88.23}\\ \rowcolor[HTML]{EFEFEF} 
    1$\times 10^{-7}$ &1 & 87.71 &89.47 &87.33 &86.40 &89.02&87.99\\ \rowcolor[HTML]{EFEFEF} 
    1$\times 1e^{-8}$ &1 & 30.50 &30.02 &30.54 &30.36 &28.77&30.04\\ \rowcolor[HTML]{EFEFEF} 
    1$\times 1e^{-10}$ &1 & 29.28 &31.38 &30.61 &31.48 &29.69&30.49\\ \rowcolor[HTML]{EFEFEF} 
    0&1&30.02&30.63&32.64&31.31&32.17&31.35\\
    \hline
    \end{tabular}%
    }
  \label{tabsy5}%
\end{table*}%

\begin{table*}[t]
  \centering
  \caption{Ablation study of proposed components on MP-78 for OVKD. Experiments are conducted on both Setting A and Setting B. PCK@0.2 is used as the metric.}
    \setlength{\tabcolsep}{4.6mm}{\begin{tabular}{ccc|ccccc|c}
    \hline
    \rowcolor[HTML]{ECF4FF}
    Baseline & DDMM & VKRA & Split1 &Split2 &Split3 &Split4& Split5& Mean (PCK@0.2 $\uparrow$) \\
    \hline \rowcolor[HTML]{EFEFEF} 
    \multicolumn{9}{c}{Setting A} \\ \rowcolor[HTML]{EFEFEF} 
    \hline
    \ding{52}& \ding{54} &\ding{54}& 42.02&44.00&42.55&43.80&42.26&42.93\\ \rowcolor[HTML]{EFEFEF} 
    \ding{52}& \ding{52} &\ding{54}&69.64&57.86&67.95&62.10&71.92&65.89\\ \rowcolor[HTML]{EFEFEF} 
    \ding{52}& \ding{52} &\ding{52} &79.02&71.35&76.68&79.79&74.67&76.30\\ \rowcolor[HTML]{EFEFEF}     
    \hline   \rowcolor[HTML]{ECF4FF}   
    \multicolumn{9}{c}{Setting B}\\ \rowcolor[HTML]{ECF4FF}  
    \hline
     \ding{52}& \ding{54} &\ding{54}&56.06&55.36&54.35&53.07&50.66&53.90 \\ \rowcolor[HTML]{ECF4FF} 
    \ding{52}& \ding{52} &\ding{54}& 72.96 & 77.66& 76.63 & 78.26 & 62.43 & 73.59 \\ \rowcolor[HTML]{ECF4FF} 
    \ding{52}& \ding{52} &\ding{52} & 84.02 & 87.99& 83.22& 83.20& 80.25& 83.74\\      
    \hline
    \end{tabular}%
    }
  \label{tabsy3}%
\end{table*}%
\begin{table*}[t]
  \centering
  \caption{Performance (PCK0.2) comparison of different attention blocks in Setting A for the OVKD task. 
}
    \setlength{\tabcolsep}{4.8mm}{\begin{tabular}{cc|ccccc|c}
    \hline
    \rowcolor[HTML]{ECF4FF}
    $\mathrm{Self}\_ \mathrm{Attention}$  & $\mathrm{Cross}\_ \mathrm{Attention}$ &  Split1 &Split2 &Split3 &Split4& Split5& Mean (PCK@0.2 $\uparrow$) \\
    \hline \rowcolor[HTML]{EFEFEF} 
    1 & 3 & 65.43 &53.78 &48.80&56.90&57.97&56.58\\ \rowcolor[HTML]{EFEFEF} 
    2 & 3 &74.89& 61.87 &69.55&78.56&70.39&71.05\\ \rowcolor[HTML]{EFEFEF} 
    3 & 3 &79.02&71.35&76.68&79.79&74.67&76.30 \\ \rowcolor[HTML]{EFEFEF} %76.30 
    4 & 3 &82.49 & 83.15 &72.00&76.66&74.33&77.73\\  \rowcolor[HTML]{EFEFEF} 
    \hline
    3 & 1 &77.04&71.16&65.19&69.18&66.65&69.84\\ \rowcolor[HTML]{EFEFEF} 
    3 & 2 &79.44& 69.07 &78.38&76.49&73.75&74.43\\   \rowcolor[HTML]{EFEFEF}   
    3 & 4 &79.62&67.00&75.69&76.41&71.71&74.09\\   \rowcolor[HTML]{EFEFEF} 
    \hline
    \end{tabular}%
    }
  \label{tabsy_new}%
\end{table*}%

\textbf{Evaluation Metrics.}
We employ the Probability of Correct Keypoint (PCK) and Normalized Mean Error (NME) metrics to assess the accuracy of keypoint detection. To mitigate category bias, we compute and present the average PCK and average NME across all dataset splits. This approach ensures a balanced and thorough evaluation of our model's performance in keypoint detection.

PCK measures the accuracy of a predicted keypoint by comparing its normalized distance to the actual ground-truth location, with respect to a predefined threshold ($\sigma$). In line with the methodologies of POMNet~\citep{xu2022pose} and CapeFormer~\citep{DBLP:conf/cvpr/0004HMH023}, we report PCK@0.2 results in our experiments, setting $\sigma$ to 0.2 for each category across all dataset splits. 
Additionally, we report PCK@0.05, where $\sigma$ is set to 0.05, demanding more precise predictions compared to $\sigma=0.02$.
NME is defined similarly to HRNet V2~\citep{wang2020deep}, where the normalization distance refers to the longest side of the ground-truth bounding box.

\begin{table*}[t]
  \centering
  \caption{Performance (PCK0.2) comparison of different $\mathrm{Text\_Encoder}$ and different $\mathrm{Vision\_Encoder}$ configurations. FLOPs represents the computational complexity of the $\mathrm{Vision\_Encoder}$. 
  (CLIP) represents the pre-trained image encoder from CLIP~\citep{radford2021learning}.} 
  % FLOPs represents the computational complexity of the $\mathrm{Vision\_Encoder}$. 
    \setlength{\tabcolsep}{3.8mm}{\begin{tabular}{cc|ccccc|c}
    \hline
    \rowcolor[HTML]{ECF4FF}
    $\mathrm{Text\_Encoder}$  & $\mathrm{Vision\_Encoder}$ (FLOPs) &  Split1 &Split2 &Split3 &Split4& Split5& Mean (PCK@0.2 $\uparrow$) \\
    \hline \rowcolor[HTML]{EFEFEF}
    \multicolumn{8}{c}{Setting A} \\ \rowcolor[HTML]{EFEFEF}
    \hline
    \textbf{\textit{Ours:} CLIP-B/32} & \textbf{ResNet50 (5.40G)} &79.02&71.35&76.68&79.79&74.67&76.30\\ \rowcolor[HTML]{EFEFEF}
    \hline
    CLIP-Res50 & ResNet50 (5.40G)& 60.60&50.34&65.61&60.07&40.89 & 55.50\\ \rowcolor[HTML]{EFEFEF}   
    CLIP-B/16 & ResNet50 (5.40G)&82.39&72.58&73.69&80.89&81.82&78.27\\ \rowcolor[HTML]{EFEFEF}    
    \hline
    CLIP-B/32 & MobileNet V2 (0.42G)  & 58.17 & 51.76 & 60.24 & 64.00& 66.58 & 60.15 \\ \rowcolor[HTML]{EFEFEF}
    CLIP-B/32 & EfficientNet-B0 (0.53G)  & 55.22 & 43.37  & 44.30 &52.53 & 39.28 & 46.94\\ \rowcolor[HTML]{EFEFEF}
    CLIP-B/32 & EfficientNet-B3 (1.33G)  & 57.45 & 46.43 & 54.95  & 57.54  & 45.87 &  52.45 \\     \rowcolor[HTML]{EFEFEF}
    \hline
    CLIP-B/32 & ViT-B/32 (CLIP) (3.83G)  &52.54&49.25&59.23&50.30&45.17&51.30\\ \rowcolor[HTML]{ECF4FF}
    \hline \rowcolor[HTML]{ECF4FF} 
    \multicolumn{8}{c}{Setting B} \\ \rowcolor[HTML]{ECF4FF} \hline
    \textbf{\textit{Ours:} CLIP-B/32} & \textbf{ResNet50 (5.40G)} &84.02 & 87.99& 83.22& 83.20& 80.25& 83.74 \\\rowcolor[HTML]{ECF4FF} 
    \hline
    CLIP-Res50 & ResNet50 (5.40G) &  77.22& 80.70  & 78.81 & 80.85 & 79.46 &  79.41\\ \rowcolor[HTML]{ECF4FF} 
    CLIP-B/16 & ResNet50 (5.40G) & 83.49 & 89.19 & 83.84 & 83.96 & 81.06 &  84.31\\    \rowcolor[HTML]{ECF4FF} 
    \hline
    CLIP-B/32 & MobileNet V2 (0.42G) & 73.03 & 65.65 & 60.42 & 59.35 & 55.68 & 62.83\\ \rowcolor[HTML]{ECF4FF}  
    CLIP-B/32 & EfficientNet-B0 (0.53G) & 78.25 & 73.49 & 75.93 &80.16 &73.26 & 76.22\\\rowcolor[HTML]{ECF4FF}
    CLIP-B/32 & EfficientNet-B3 (1.33G) & 80.28 & 87.14 & 79.86 & 82.07&75.18 &80.71\\\rowcolor[HTML]{ECF4FF} 
    \hline
    CLIP-B/32 & ViT-B/32 (CLIP) (3.83G)  & 70.74 & 73.90  & 60.68  & 73.75 & 61.55 & 68.12\\     
    \hline
    \end{tabular}%
    }
  \label{tabsy4}%
\end{table*}%

\begin{table*}[t]
  \centering
  \caption{Performance (PCK@0.2) of KDSM on different super-categories in Setting A for the OVKD task.
}
    \setlength{\tabcolsep}{6.7mm}{\begin{tabular}{c|ccccc|c}
    \hline
    \rowcolor[HTML]{ECF4FF} 
    Super-Category &  Split1 &Split2 &Split3 &Split4& Split5& Mean (PCK@0.2 $\uparrow$) \\ \rowcolor[HTML]{EFEFEF}
    \hline
    Face & 85.05 &77.56 &83.52&87.31&76.01&81.89 \\ \rowcolor[HTML]{EFEFEF}
    Body & 76.73&68.61&73.67&76.37&74.49&73.97\\  \rowcolor[HTML]{EFEFEF}
    Face w/ Body&79.02&71.35&76.68&79.79&74.67&76.30\\
    \hline
    \end{tabular}%
    }
  \label{tabsy6}%
\end{table*}%

\subsection{Implementation Details}
In our setup, the default $\mathrm{Vision\_Encoder}$ is ResNet50~\citep{he2016deep}, pre-trained on the ImageNet dataset~\citep{deng2009imagenet} by default unless otherwise specified. The $\mathrm{Self\_Attn.}$ module consists of three layers, each featuring a multi-head self-attention mechanism and a feed-forward neural network (FFN). This self-attention component is equipped with four attention heads and an embedding dimension of 512, with a dropout rate set at 0.1. The FFN includes two fully connected layers, an embedding dimension of 512, and 2048 feedforward channels. We employ ReLU as the activation function and maintain a dropout rate of 0.1.
The $\mathrm{Cross\_Attn.}$ component also comprises three layers. Each layer incorporates a multi-head self-attention mechanism, a multi-head cross-attention mechanism, and an FFN. The FFN configuration mirrors that of the $\mathrm{Self\_Attn.}$
For text encoding, we default to using CLIP~\citep{radford2021learning}'s $\mathrm{Text\_Encoder}$, pre-trained alongside the ViT-B/32 $\mathrm{Vision\_Encoder}$ on image-text paired data, unless an alternative specification is provided.

The objects of interest are extracted using their bounding boxes and resized to dimensions of $256\times256$. To bolster the model's generalization capabilities, data augmentation techniques such as random scaling (varying from $-15\%$ to $15\%$) and random rotation (varying from $-15^{\circ}$ to $15^{\circ}$) are applied. Training is carried out across 4 GPUs, each with a batch size of 64, for a total of 210 epochs.

\subsection{Results for OVKD}

\textbf{Setting A: Diverse Keypoint Categories.} Table~\ref{tabsy1} presents the performance comparison between the baseline framework and KDSM on the MP-78 dataset for this setting. The table highlights that KDSM consistently surpasses the baseline in all five dataset splits.
The quantitative comparison of the results shows a significant performance improvement when using the KDSM framework. 
The Mean (PCK@0.2) score across all five splits increases from 42.93 for the baseline to 88.23 for the KDSM framework, resulting in a remarkable enhancement of 45.30 points. Similarly, the Mean (PCK@0.05) score increases from 11.93 for the baseline to 62.35 for the KDSM framework, indicating a substantial enhancement of 50.42 points. In addition, the Mean NME decreases from 29.72 for the baseline to 7.93 for the KDSM framework, showcasing an improvement of 21.79 points.
This indicates that the KDSM approach is more effective at handling the ``Diverse Keypoint Categories" setting in the zero-shot fashion.
The superior performance of the KDSM framework on the ``Diverse Keypoint Categories" setting can be attributed to its capacity to better align and match semantic information from text prompts with local visual features, as well as its ability to effectively transfer knowledge to unseen $({\textit{animal species}}, {\textit{keypoint category}})$ pairs.

\textbf{Setting B: Varied Animal Species.} Table~\ref{tabsy2}\footnote{We refer to the method ``\textbf{F}ew-\textbf{s}hot keypoint detection with \textbf{u}ncertainty \textbf{l}earning for \textbf{u}nseen \textbf{s}pecies" as FS-ULUS.} displays the performance comparison between the baseline framework and KDSM on the MP-78 dataset for the ``Varied Animal Species" setting under a zero-shot setting. Additionally, it compares the results with class-agnostic keypoint detection methods under 1-shot and 5-shot settings.

The KDSM framework significantly outperforms the baseline in the zero-shot setting, demonstrating its effectiveness in handling unseen animal species without category-specific training. The enhanced performance of the KDSM framework is due to the efficient knowledge transfer from seen to unseen $\{\textit{animal species}\}$.
Besides, recent research~\citep{xu2022pose,DBLP:conf/cvpr/0004HMH023} has developed few-shot species class-agnostic keypoint detection techniques that can identify keypoints across various animal species without category-specific training. However, these techniques typically rely on support images with annotations during both the training and testing phases.
In contrast, our OVKD approach using the KDSM framework does not require support images by leveraging text prompts $\{\textit{animal species}\}$ and $\{\textit{keypoint category}\}$ for semantic guidance. 

OVKD and few-shot species class-agnostic keypoint detection represent distinct methodological approaches, making direct comparisons challenging, so we primarily benchmark against our baseline. However, we also highlight the performance gap contrast with few-shot species class-agnostic keypoint detection methods at a macro level.
Our method demonstrates comparable results to these few-shot species class-agnostic keypoint detection approaches and outperforms the state-of-the-art 1-shot solution, CapeFormer~\citep{DBLP:conf/cvpr/0004HMH023}, across all three metrics. This emphasizes the effectiveness of our approach.
Furthermore, our zero-shot OVKD even surpasses the 5-shot setting of FS-ULUS~\citep{lu2022few}, MAML~\citep{finn2017model}, Fine-tune~\citep{nakamura2019revisiting}, and POMNet~\citep{xu2022pose} across all three metrics. It should be noted that methods like POMNet and CapeFormer have limitations during training as they cannot access images of new categories and rely on support images during testing. Hence, it is reasonable for our zero-shot method to exhibit superior performance compared to few-shot solutions.
In particular, when considering the PCK@0.05 metric, we outperform the state-of-the-art CapeFormer by 9.25 points (56.20 vs. 46.95). It is worth noting that PCK@0.05 requires more precise predictions of keypoint locations compared to the less stringent PCK@0.2 metric. By evaluating keypoint detection performance using different metrics such as PCK and NME, we provide a comprehensive analysis of our method's performance.

\begin{figure}
\centering
  \includegraphics[width=0.45\textwidth]{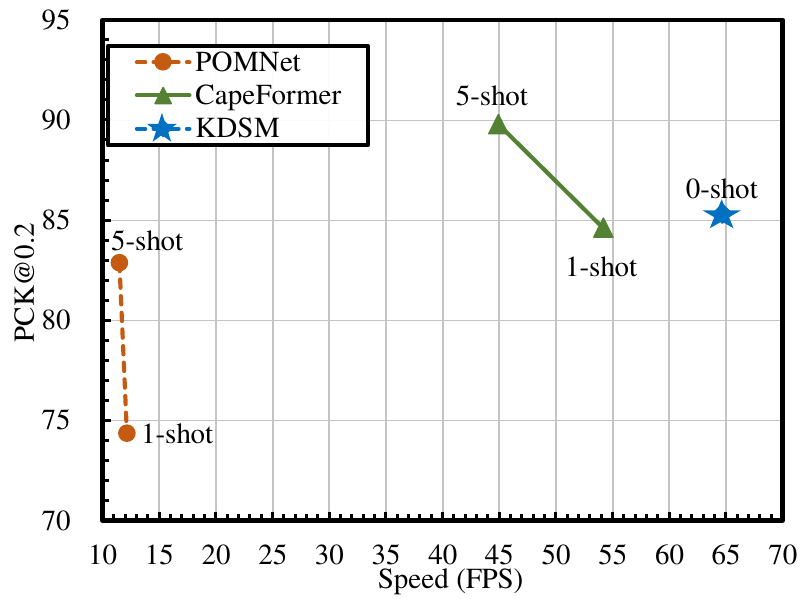}  
  \caption{Comparison of the trade-off between PCK@0.2 and Speed for Setting B. The speed is measured using Frames Per Second (FPS) on a single NVIDIA A100-SXM-80GB card. The test is conducted using an average of 1000 images for one species.
  }  
  \label{fig:speed}
\end{figure}

\begin{figure}
\centering
  \includegraphics[width=0.45\textwidth]{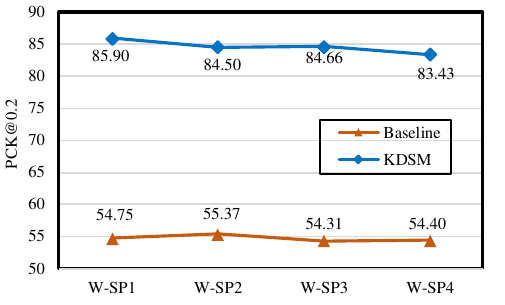}  
  \caption{Comparisons of the performance (PCK@0.2) between the baseline and KSDM on long-tail species for the AnimalWeb dataset.
  }  
  \label{fig:longtail}
\end{figure}

\textbf{Inference Speed.} 
In Fig.~\ref{fig:speed}, we compare the trade-off between PCK@0.2 and Inference Speed (Frames Per Second) with state-of-the-art few-shot solutions, namely POMNet\citep{xu2022pose} and CapeFormer~\citep{DBLP:conf/cvpr/0004HMH023}. The speed is reported as an average of 1000 test images.  As shown in the figure, it is evident that our KDSM method surpasses POMNet~\citep{xu2022pose} in both average speed and accuracy. Furthermore, our approach exhibits a significant speed advantage compared to CapeFormer~\citep{DBLP:conf/cvpr/0004HMH023}. These findings highlight the promising prospects of our method for practical applications.

\textbf{Long-tail Animal Species.} AnimalWeb~\citep{khan2020animalweb} is a long-tail keypoint detection dataset consisting of 350 different animal species. The number of annotated images per species ranges from 1 to 239, reflecting the varying difficulty in data collection and substantial species imbalance.  
We prepare the data of uncommon animal species from AnimalWeb by first excluding the categories that are already present in MP-78. Then, we sort the remaining species on the AnimalWeb dataset based on the number of annotated samples. 280 species with relatively smaller number of samples are evenly divided into four partitions, denoted as W-SP1, W-SP2, W-SP3, and W-SP4, and each partition contains 70 species. Notably, W-SP1 consists of the most common species, while W-SP4 represents the long-tail species with the fewest annotations. We evaluate the baseline and KDSM in each partition using the five models trained in setting B, and report the average PCK@0.2 across the five models as the final result for each method, respectively.
As shown in Fig.~\ref{fig:longtail}, even on the long-tail species set W-SP4, KDSM achieves a PCK@0.2 score of 83.43, which is comparable with the relatively common species set W-SP1. This demonstrates the robustness of KDSM in handling long-tail categories. We also observe that KDSM gets a slightly lower result for W-SP2 compared to W-SP3 (84.50 vs. 84.66), which is expected due to inherent differences between species, including variations in pose and other factors, indicating that the detection accuracy is not solely determined by species prevalence.
Furthermore, similar to the results shown in Table~\ref{tabsy2}, there is a noticeable performance gap between the baseline and KDSM, indicating the superiority of KDSM in the OVKD task.

\begin{figure*}
\centering
  \includegraphics[width=0.9\textwidth]{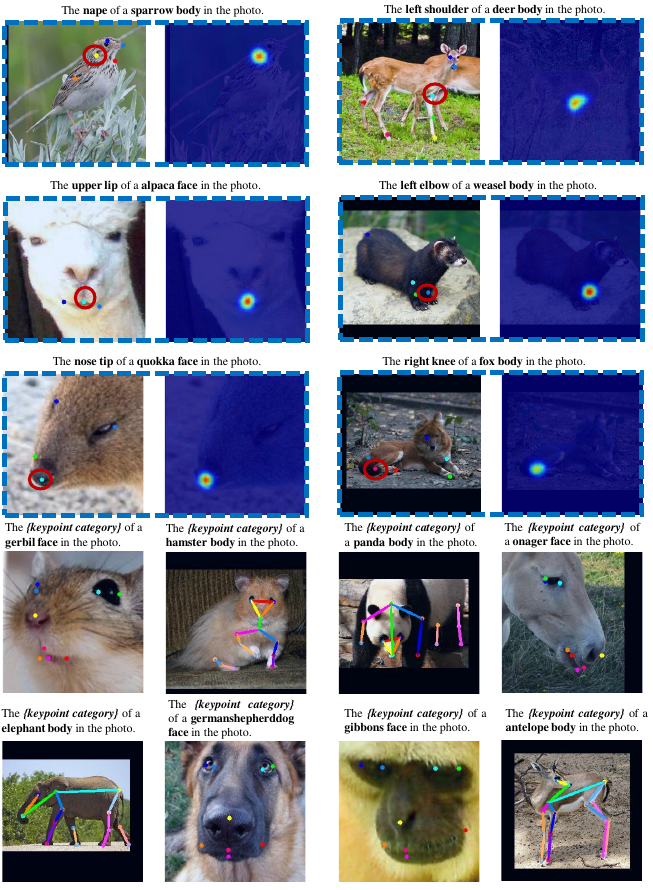}  
  \caption{Visual results of KDSM on the test sets of two experiment settings of OVKD. The first three rows show the heatmaps for Setting A, and the last two rows show the results for Setting B. KDSM achieves satisfactory results in both two settings. Due to space limitations, we use $\{\textit{keypoint category}\}$ to represent the keypoint categories.
  }  
  \label{fig:qua}
\end{figure*}
\begin{figure*}
\centering
  \includegraphics[width=0.8\textwidth]{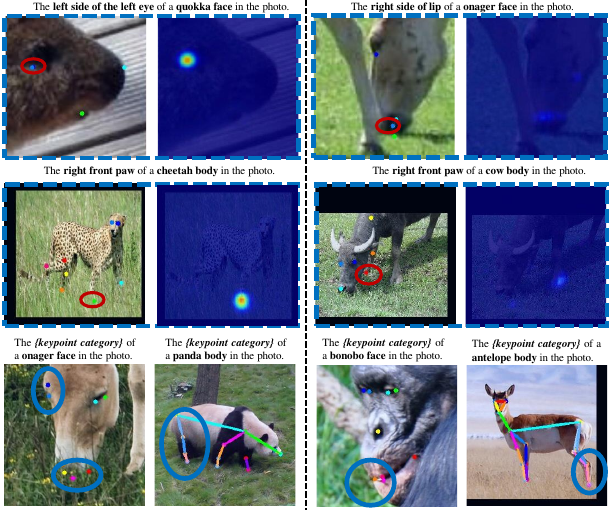}  
  \\
  {\footnotesize \qquad\qquad\qquad\qquad\qquad\qquad\quad\quad\qquad(a)\hspace*{\fill}\qquad\qquad\qquad\qquad\quad(b)\hspace*{\fill}}
  \caption{Visual results of challenging KDSM on the test sets of two experiment settings of OVKD. 
  (a) Demonstrates that KDSM can handle challenging scenarios involving body occlusion, environmental occlusion, and complex poses.
(b) Illustrates the failure cases of KDSM in challenging keypoint detection.
  The points circled in red represent the ground-truth keypoint locations corresponding to the heatmaps.
  The blue circles enclose the challenging regions of keypoint detection.
  Due to space limitations, we use $\{\textit{keypoint category}\}$ to represent the keypoint categories.
  }  
  \label{fig:sup}
\end{figure*}

\subsection{Ablation Study}
In this section, we do some ablation experiments about the hyperparameter settings of the loss function, domain distribution matrix matching, VKRA module, $\mathrm{Vision\_Encoder}$ and $\mathrm{Text\_Encoder}$. The default setting is $\alpha=1e^{-3}$ and $\beta=1$.

\textbf{Discussion of the Loss Function of KDSM.}
We explore various hyperparameter configurations in this section. Table~\ref{tabsy5} illustrates how these settings impact KDSM's performance in the OVKD Setting A evaluation. 
We observe that as the value of $\alpha$ is reduced from 1 to $10^{-10}$, while maintaining $\beta$ at a constant 1, the Mean (PCK@0.2) shows an increasing trend. The optimal performance is attained at $\alpha = 10^{-6}$, resulting in a Mean (PCK@0.2) of 88.23. Conversely, further reducing $\alpha$ below $10^{-6}$, or setting it to 0, leads to a decrease in Mean (PCK@0.2), suggesting an ideal range for $\alpha$'s value. Notably, when $\alpha$ is set to 0, the Mean (PCK@0.2) falls sharply to 31.15, underscoring the significance of domain distribution matrix matching.

\textbf{Domain Distribution Matrix Matching.}
Table~\ref{tabsy3} demonstrates a significant improvement in Mean (PCK@0.2) scores with the inclusion of DDMM. In setting A, the Mean (PCK@0.2) is enhanced from 42.93 to 65.89, while in setting B, Mean (PCK@0.2) is elevated from 53.90 to 73.59. This substantial increase attests to DDMM's effectiveness in promoting knowledge transfer between seen and unseen keypoint categories. Moreover, the uniform improvement across all dataset splits underscores the robustness and adaptability of our proposed method, emphasizing its suitability for diverse real-world applications.

\textbf{Vision-Keypoint relational Awareness Module.}
Table~\ref{tabsy3} shows that integrating the baseline framework with both DDMM and VKRA Module leads to a notable increase in Mean (PCK@0.2) scores. Specifically, the Mean (PCK@0.2) rises from 42.93 in the baseline without these components to 76.30 in setting A and from 53.90 to 83.74 in setting B when incorporating both DDMM and VKRA modules. This improvement underscores the critical necessity of the VKRA module in our methodology, as it adeptly discerns the semantic connections between visual features and text prompts, thereby enhancing generalization capabilities for unseen keypoint categories.

\textbf{Attention Layers in Vision-Keypoint relational Awareness Module.}
Our study also delves into the optimal number of self-attention and cross-attention layers within the VKRA module. The findings, as depicted in Table~\ref{tabsy_new}, indicate that augmenting the number of self-attention blocks from 1 to 3 leads to a marked improvement in performance (compare row 1 with row 3). However, adding a fourth self-attention block doesn't contribute substantially to further gains (compare row 3 with row 4). A similar pattern is observed with the number of cross-attention blocks, leading us to implement three cross-attention blocks in our final configuration.

\textbf{Discussion on the Choice of Vision Encoder.}
Following previous research~\citep{ni2022expanding}, we deviate from using the frozen CLIP visual encoder and instead train a task-specific visual encoder, but we still leverage the language model's knowledge (that is why we can achieve OVKD). 
The results of deploying various Vision Encoders such as MobileNet V2~\citep{sandler2018mobilenetv2}, EfficientNet-B0 and B3~\citep{tan2019efficientnet}, as well as ResNet50~\citep{he2016deep} within the KDSM are detailed in Table~\ref{tabsy4}.
Even if the extremely lightweight models such as MobileNet V2, EfficientNet-B0 and B3, KDSM still achieves reasonable performance and outperforms the OVKD baseline method (42.93 PCK@0.2 for setting A and 53.90 PCK@0.2 for setting B).
We choose ResNet in our implementation in order to ensure a fair comparison with state-of-the-art few-shot keypoint detectors, i.e., POMNet and CapeFormer that use ResNet50 as vision encoder.
Besides, our attempt to utilize the CLIP pre-trained Vision Transformer~\citep{dosovitskiy2020image} (ViT-B/32) falls short of expectations. This could be attributed to the fact that OVKD necessitates precise joint localization and detailed region-level feature extraction to handle diverse pose variations, which contrasts with the global image-level features captured by the CLIP visual encoder.

\textbf{Discussion on the Choice of Text Encoder.}
Table~\ref{tabsy4} compares different $\mathrm{Text\_Encoders}$ that have been pre-trained in conjunction with distinct image encoders of CLIP~\citep{radford2021learning}. In setting A, the Mean (PCK@0.2) scores are 55.50, 76.30, and 78.27 for the $\mathrm{Text\_Encoders}$ pre-trained with ResNet50, ViT-B/32, and ViT-B/16 image encoders, respectively. In setting B, the Mean (PCK@0.2) scores are 79.41, 83.74, and 84.31 for the $\mathrm{Text\_Encoders}$ pre-trained with ResNet50, ViT-B/32, and ViT-B/16 image encoders, respectively. Notably, the $\mathrm{Text\_Encoder}$ corresponding to ViT-B/16 image encoder achieves the highest performance.
The performance disparity among the  $\mathrm{Text\_Encoders}$ indicates that using a more robust $\mathrm{Text\_Encoder}$, particularly one pre-trained with a more powerful image encoder, leads to improved results. Although we utilize the $\mathrm{Text\_Encoder}$ pre-trained with ViT-B/32 image encoder in this study, this finding highlights the significant potential for enhancing our method's performance by integrating a stronger $\mathrm{Text\_Encoders}$.

\textbf{Discussion of OVKD Task for Different Super-Categories.}
To assess KDSM's capability in managing various super-categories within the OVKD task, we segregated the MP-78 dataset into two distinct, non-overlapping super-categories: Face and Body. Table~\ref{tabsy6} demonstrates KDSM's differing performance in these categories. Specifically, it achieved Mean (PCK@0.2) scores of 81.89 for the Face category and 73.97 for the Body category, clearly showing a superior performance in the Face category. The comparatively lower score for the Body category likely stems from the more complex and varied body poses. Despite the strong results, there appears to be potential for further enhancement, particularly in the Body category's performance.

\subsection{Qualitative Results}

In Fig.~\ref{fig:qua}, we showcase the performance of KDSM in two experimental scenarios of OVKD. The top three rows depict the heatmaps for novel keypoint categories in setting A, and the bottom two rows display the actual keypoint detection outcomes in setting B. These visualizations effectively highlight KDSM's capability to adeptly navigate the OVKD task in both experimental setups.

\section{Future work}

Firstly, our research focuses on achieving OVKD, a new and promising research topic, with satisfactory performance on regular scenes. Further improvement in challenging scenarios (e.g., occlusion, lighting, and resolution) will be left for our future work. Unlike traditional methods that rely on manual annotation, OVKD offers valuable recognition to arbitrary keypoints without prior annotation. 
We include some results of our method's performance in occlusion scenarios in Fig.~\ref{fig:sup}(a), demonstrating its capability to handle certain occlusion cases effectively. However, we also present some instances where our method encounters challenges under occlusion, as seen in Fig.~\ref{fig:sup}(b), indicating areas for potential improvement.

Secondly, we notice certain issues with individual predicted heatmaps in Fig.~\ref{fig:qua}, such as ``The left shoulder of a deer body in the photo," exhibiting the problem of ``anisotropic Gaussian distribution". In future work, we can try to find appropriate methods to address the ``anisotropic Gaussian" issue in the OVKD task by adjusting the loss function like LUVLi~\citep{kumar2020luvli} and STAR Loss~\citep{zhou2023star}.

Last but not least, we plan to explore a new research direction that employs a hybrid approach utilizing both textual and visual prompts in the future. This new direction can leverage visual prompts to detect keypoints in the absence of specific semantic information.
For instance, the datasets, such as WFLW~\citep{wu2018look} (98 annotated keypoint categories) and CatFLW~\citep{martvel2023catflw} (48 annotated keypoint categories), are annotated with a considerable number of non-semantic keypoint categories, which will be effectively addressed through this new research direction.

\section{Conclusion}

We address the challenges inherent in traditional image-based keypoint detection methods for animal (including human) body and facial keypoint detection by introducing the \textbf{O}pen-\textbf{V}ocabulary \textbf{K}eypoint \textbf{D}etection (OVKD) task. This task is designed to identify keypoints in images, regardless of whether the specific animal species and keypoint category have been encountered during training. Our novel framework, Open-Vocabulary \textbf{K}eypoint \textbf{D}etection with \textbf{S}emantic-feature \textbf{M}atching (KDSM), leverages the synergy of advanced language models to effectively bridge the gap between text and visual keypoint features. KDSM integrates innovative strategies such as \textbf{D}omain \textbf{D}istribution \textbf{M}atrix \textbf{M}atching (DDMM) and other special modules, such as the  \textbf{V}ision-\textbf{K}eypoint \textbf{R}elational \textbf{A}wareness (VKRA) module, leading to significant performance enhancements. Specifically, we observed a 45.30-point improvement in detecting diverse keypoint categories and a 31.43-point improvement for varied animal species compared to the baseline framework. Notably, KDSM achieves comparable results with those of state-of-the-art few-shot species class-agnostic keypoint detection methods. The proposed approach lays the groundwork for future exploration and advancements in OVKD, driving further improvements in quantitative performance metrics.

\section*{Declarations}

\noindent \textbf{Data Availability} The dataset MP-100 for this study can be downloaded at: \url{ https://github.com/luminxu/Pose-for-Everything}. Our reorganized and partitioned dataset MP-78 is released together with our source code.

\noindent \textbf{Conflict of interest} The authors declare that they have no conflict of interest.

% Non-BibTeX users please use
{\footnotesize
	\bibliographystyle{plainnat}
	\bibliography{egbib}
}
% \end{flushend}
\balance
\end{document}